%% file: main.tex
\documentclass[10pt,twocolumn,letterpaper]{article}
\newcommand{\minisection}[1]{\noindent {\bf #1}.}
\PassOptionsToPackage{table}{xcolor}
\usepackage[pagenumbers]{wacv}
\usepackage[margin=1in]{geometry}
\usepackage{array} % for p{width} column types
\usepackage{arydshln} % for dashed lines

\newcommand{\nomemanipulation}{Leave-one-out}

\definecolor{wacvblue}{rgb}{0.21,0.49,0.74}
\usepackage[pagebackref,breaklinks,colorlinks,allcolors=wacvblue]{hyperref}

\definecolor{tabhighlight}{rgb}{0.88,0.95,1}

\def\wacvPaperID{2161} % *** Enter the WACV Paper ID here
\def\confName{WACV}
\def\confYear{2026}

\title{No MoCap Needed: Post-Training Motion Diffusion Models with Reinforcement Learning using Only Textual Prompts}
\author{Macaluso Girolamo$^{*}$\\
University of Florence\\
{\tt\small girolamo.macaluso@unifi.it}
\and
Mandelli Lorenzo$^{*}$\\
University of Florence\\
{\tt\small lorenzo.mandelli@unifi.it}
\and
Mirko Bicchierai\\
University of Florence\\
{\tt\small mirko.bicchierai@unifi.it}
\and
Stefano Berretti\\
University of Florence\\
{\tt\small stefano.berrettin@unifi.it}
\and
Andrew D. Bagdanov\\
University of Florence\\
{\tt\small andrew.bagdanov@unifi.it}
}

\begin{document}
\maketitle
\def\thefootnote{*}\footnotetext{Equal contribution}
\def\thefootnote{\arabic{footnote}}
\begin{abstract}

Diffusion models have recently advanced human motion generation, producing realistic and diverse animations from textual prompts. However, adapting these models to unseen actions or styles typically requires additional motion capture data and full retraining, which is costly and difficult to scale. We propose a post-training framework based on Reinforcement Learning that fine-tunes pretrained motion diffusion models using only textual prompts, without requiring any motion ground truth. Our approach employs a pretrained text–motion retrieval network as a reward signal and optimizes the diffusion policy with Denoising Diffusion Policy Optimization, effectively shifting the model’s generative distribution toward the target domain without relying on paired motion data. We evaluate our method on cross-dataset adaptation and leave-one-out motion experiments using the HumanML3D and KIT-ML datasets across both latent- and joint-space diffusion architectures. Results from quantitative metrics and user studies show that our approach consistently improves the quality and diversity of generated motions, while preserving performance on the original distribution. Our approach is a flexible, data-efficient, and privacy-preserving solution for motion adaptation.

\end{abstract}

\section{Introduction}
Human motion generation is a foundational component of diverse applications, spanning Computer Animation, Virtual and Augmented Reality, Human–Computer Interaction, and Robotics~\cite{holden2016deep, argall2009survey}. By synthesizing realistic and semantically rich movements, generative motion models can simplify content creation, enhance immersion, and enable dynamic, natural user experiences.

Recent breakthroughs in generative modeling, particularly denoising diffusion probabilistic models~\cite{tevet2022humanmotiondiffusionmodel, motion_diffuse, conditioned_music_0, conditioned_music_1, latent_diffusion}, have elevated the quality and fidelity of synthesized human motion. Leveraging multi-modal conditioning, diffusion-based approaches can translate high-level instructions, such as textual descriptions, into continuous, lifelike animations~\cite{ho2020denoising, Guo_2022_CVPR, tevet2022humanmotiondiffusionmodel}.

However, a key limitation of existing motion diffusion models (DMs) lies in their lack of adaptability. As shown by~\cite{ma2023motiontransfer, liu2024cross}, even minor shifts in motion distribution, such as domain changes or novel styles, can lead to severe performance degradation, with FID scores often doubling or tripling on out-of-domain evaluations. This issue is particularly pronounced for Human Motion DMs, largely due to the relatively small size of publicly available datasets. Current models struggle to generalize in a zero-shot manner to unseen actions or motion styles, and adapting them typically requires additional ground-truth motion capture data along with %lengthy 
retraining, a process that is costly, labor-intensive, and time-consuming. These constraints significantly hinder the adaptability and practical deployment of diffusion-based motion generators in novel or specialized application domains.

In contrast, the image generation community has made significant progress in post-training alignment methods, in particular Reinforcement Learning (RL) based fine-tuning, that shift pre-trained DMs toward new distributions. By optimizing a model with task-specific reward functions (e.g., perceptual scores, or aesthetic quality), these techniques shift the generative distribution in a desired direction, allowing rapid adaptation to novel concepts, while enhancing output quality~\cite{black2023training, fan2023dpok, clark2023draft, zhang2024largescalereinforcementlearningdiffusion, li2024textcraftortextencoderimage, wei2024powerfulflexiblepersonalizedtexttoimage}. However, directly applying these methods to motion generation presents unique challenges: motion data is inherently temporal and high-dimensional, motion-text alignment is more complex than image-text relationships, and suitable reward functions for motion quality assessment are less established.

%Inspired by these advances, 
In this paper, we introduce an RL-based post-training framework for pretrained human motion DMs, allowing them to specialize in new motion categories or stylistic domains. Unlike existing motion adaptation approaches~\cite{han2025reindiffuse,mao2024learning,liu2024motionrl}, \emph{our method does not require additional motion capture data}. Instead, it leverages a pre-trained text-motion alignment network as the sole reward signal, specifically using a Text-Motion Retrieval (TMR)~\cite{tmr} model that provides semantic alignment scores between generated motions and textual descriptions. This ground truth-free design inherently preserves privacy: when motion datasets are proprietary or privacy restricted, as is common when acquisition is costly, developers can share a trained evaluator without releasing the raw data, thus allowing knowledge transfer without compromising confidentiality.

To assess our framework, we conduct experiments across challenging scenarios: cross-dataset experiments in which a model pre-trained on one motion dataset is adapted to a second, unseen dataset using only textual prompts; \textit{\nomemanipulation~class}, where a model is trained with one action category removed (e.g., object manipulations) and then fine-tuned on the excluded class using only the corresponding prompts. 
In addition, we evaluate our method on both latent diffusion and joint-space motion generation models, as well as across different motion representations, to assess its generality. 
The results demonstrate that RL-based post-training consistently improves both the quality and semantic alignment of generated motions, underscoring its potential as a flexible, data-efficient, and privacy-conscious approach for real-world motion synthesis.
Our contributions are summarized as follows:
\begin{itemize}
    \item We introduce an RL-based fine-tuning pipeline that effectively generalizes human motion DMs to new datasets and previously unseen motion categories. This is achieved without requiring any ground-truth motion capture data, but using only textual prompts and a pre-trained text-motion retrieval model as reward signal;
    \item We demonstrate the effectiveness of our approach through comprehensive experiments involving cross-dataset fine-tuning 
    and intra-dataset fine-tuning on excluded motion categories.  

Results confirm significant improvements in zero-shot motion generation quality, with consistent gains across different experimental settings and model architectures.
\end{itemize}

\section{Related Work}
Here, we review existing approaches to human motion generation, with a focus on DMs and recent advances in RL for post-training alignment and generalization.

\smallskip

\minisection{Human Motion Generation}
Human motion synthesis is a core research area in animation, robotics, and virtual environments. Classical approaches relied on motion graphs, parametric models, or handcrafted rules to generate plausible trajectories~\cite{ kovar2002motion}. With the advent of deep learning, data-driven models such as RNNs~\cite{fragkiadaki2015recurrentnetworkmodelshuman}, GANs~\cite{Gan_0, Gan_1, kundu2018bihmpganbidirectional3dhuman}, and VAEs~\cite{VAE_0, temos} became dominant for capturing temporal dynamics and producing natural motion sequences. However, these methods struggle with diversity and controllability, particularly when generalizing to unseen motions or textual inputs.

\minisection{Diffusion Models for Human Motion Generation}
Recent works have demonstrated that denoising diffusion probabilistic models (DDPMs) are especially well-suited for human motion generation due to their ability to model complex, stochastic trajectories. Zhang \etal~\cite{motion_diffuse} introduced a diffusion-based framework for text-conditioned generation. 
Follow-up methods such as FLAME~\cite{kim2023flamefreeformlanguagebasedmotion}, MDM~\cite{tevet2022humanmotiondiffusionmodel} and ReMoDiffuse~\cite{remodiffuse} improved the fidelity and semantic alignment of generated motions. Latent DMs, including MLD~\cite{latent_diffusion} and  StableMoFusion~\cite{kim2022stablemofusion}, further reduced computational costs while retaining quality. Despite their success, these models require new data and extensive retraining when adapting to new motion types or domains.

\minisection{RL for Post-training Alignment}
The image generation community has recently embraced post-training alignment strategies based on RL. Methods like DPPO~\cite{black2023training}, DPOK~\cite{fan2023dpok}, and others~\cite{li2024textcraftortextencoderimage,wei2024powerfulflexiblepersonalizedtexttoimage} leverage reward-based optimization to align pretrained DMs with user preferences, aesthetic goals, or task-specific criteria without requiring paired supervision. These methods modify the generative process to better satisfy desired constraints, enabling flexible adaptation with minimal overhead. 

\minisection{Human Motion Models and RL}
Reinforcement learning has been extensively applied to control and imitation learning in the context of motion generation~\cite{peng2018deepmimic, merel2018neural, reinforce_1, reinforce_2, reinforce_3}, particularly to train policies that imitate expert demonstrations or optimize physical realism. 
More recently, RL has also been explored as a tool to fine-tune generative motion models. However, all of these approaches still depend on access to ground-truth motion data, which limits their flexibility and scalability.
ReinDiffuse~\cite{han2025reindiffuse} focuses on reducing physical artifacts by incorporating a reward function that encourages physically plausible motion. 
InstructMotion~\cite{mao2024learning} proposes a framework for instruction-guided human motion generation using an autoregressive transformer. 
MotionRL~\cite{liu2024motionrl} develops a VQ-VAE~\cite{VQ_VAE, motiongpt} built upon the MoMask architecture~\cite{momask}, and balances reward signals from multiple sources, including ground-truth motion, human preferences, and text adherence.

\smallskip

In contrast to these works, we propose an RL-based fine-tuning pipeline for pretrained DMs that does not require any additional motion-capture and instead leverages only reward signals derived from pretrained evaluators or heuristic objectives.
\input{Figures/teaser}

\section{Preliminaries}
Here we provide an introduction to human motion DMs and RL, which form the foundation of our approach.

\smallskip

\minisection{Diffusion Models for Human Motion Generation}
Diffusion models have recently emerged as a powerful class of generative models for human motion synthesis~\cite{motion_diffuse, tevet2022humanmotiondiffusionmodel, mandelli, petrovich24stmc, zhang2023generating}. These models learn to generate realistic motion sequences by gradually denoising a sample drawn from a known noise distribution through a learned reverse process.
Let $\mathbf{x}_0$ denote a motion sequence (e.g., a sequence of joint positions or rotations), and let $q(\mathbf{x}_t \mid \mathbf{x}_0)$ represent a predefined forward noising process that progressively adds Gaussian noise to the data over $T$ steps:
\begin{equation}
q(\mathbf{x}_t \mid \mathbf{x}_0) = \mathcal{N}(\mathbf{x}_t; \sqrt{\alpha_t} \mathbf{x}_0, (1 - \alpha_t) \mathbf{I}),
\end{equation}

\noindent
where $\{\alpha_t\}_{t=1}^T$ is a variance schedule.

The generative model learns a reverse process $p_\theta(\mathbf{x}_{t-1} \mid \mathbf{x}_t, \mathbf{c})$ that reconstructs clean motion sequences conditioned on input context $\mathbf{c}$ (e.g., textual descriptions):
\begin{equation}
p_\theta(\mathbf{x}_{t-1} \mid \mathbf{x}_t, \mathbf{c}) = \mathcal{N}\left(\mathbf{x}_{t-1}; \boldsymbol{\mu}_\theta(\mathbf{x}_t, t, \mathbf{c}), \boldsymbol{\Sigma}_t\right),
\end{equation}

\noindent
where $\boldsymbol{\mu}_\theta$ is a neural network (often a U-Net or a transformer-based denoiser) trained to predict the noise or the original signal. During sampling, a motion sequence is generated by starting from $\mathbf{x}_T \sim \mathcal{N}(0, \mathbf{I})$ and recursively applying the learned reverse process until reaching $\mathbf{x}_0$, the final denoised motion.

\smallskip

\minisection{Reinforcement Learning} RL provides a general framework for learning policies that maximize a reward signal within a Markov Decision Process (MDP)~\cite{sutton1998reinforcement}. While RL has been traditionally applied to sequential decision-making problems, recent work has shown its effectiveness for fine-tuning generative models by treating the generation process as an MDP. 
In autoregressive transformer models for motion generation, the MDP can be naturally defined over animation time steps, since the model predicts one frame at a time. 

In contrast, DMs generate the entire animation simultaneously via a series of denoising steps. Therefore, the MDP must be defined over the diffusion time steps, where each step refines the noisy sample toward a clean motion sequence. This formulation presents unique challenges: the action space is high-dimensional (the entire motion sequence), and the final output emerges only after many denoising steps, making credit assignment non-trivial.

We adopt the MDP formulation proposed by \citet{black2023training}, which has been successfully applied in the image domain, and adapt it to motion generation:
\begin{align}
    \mathbf{s}_t \triangleq (\mathbf{c}, t, \mathbf{x}_t), \quad \mathbf{a}_t \triangleq \mathbf{x}_{t-1} ,
\\
    \pi(\mathbf{a}_t \mid \mathbf{s}_t) \triangleq p_\theta(\mathbf{x}_{t-1} \mid \mathbf{x}_t, \mathbf{c}) ,
\\
   R(\mathbf{s}_t, \mathbf{a}_t) \triangleq     \begin{cases}
r(\mathbf{x}_0, \mathbf{c}) & \text{if } t = 0 \\
0 & \text{otherwise} .
\end{cases}
\end{align}
Here $\mathbf{s}_t$ is the state at diffusion step $t$, consisting of the conditioning input $\mathbf{c}$, the current timestep $t$, and the noisy animation $\mathbf{x}_t$. The action $\mathbf{a}_t$ corresponds to the denoised sample $\mathbf{x}_{t-1}$. The policy $\pi$ is defined by the DM itself, which predicts the next denoised frame. Importantly, this formulation treats the DM parameters $\theta$ as the policy parameters to be optimized. The reward is sparse and only provided at timestep $t = 0$, i.e., when the final denoised animation $\mathbf{x}_0$ is available.

\section{Method}
In this section, we describe our approach for adapting human motion generation models to new motion categories through RL post-training without relying on ground-truth motion data. It consists of three key components: policy optimization using Denoising Diffusion Policy Optimization (DDPO) with importance sampling (\S\ref{sec:ddpo}), a reward model based on text-motion retrieval (\S\ref{sec:Reward}), and efficiency improvements (\S\ref{sec:EfficientLearning}).

\subsection{Policy Optimization with DDPO} \label{sec:ddpo}
To optimize the policy represented by the denoising DM, we adopt the \textit{Denoising Diffusion Policy Optimization} (DDPO)~\cite{black2023training}. DDPO frames the reverse diffusion process as a multi-step MDP, where each denoising step corresponds to an action taken by the policy. 

To improve sample efficiency and enable multiple policy updates per batch of generated data, we use the \textit{importance sampling}~\cite{kakade2002approximately} variant of DDPO. This variant allows reweighting of old trajectories using their likelihood under the updated policy, enabling us to reuse previously collected samples for multiple training iterations. In practice, we implement this optimization using the clipped surrogate objective from \textit{Proximal Policy Optimization} (PPO)~\cite{schulman2017proximal}, which ensures stable updates by weighting the deviation between the new and old policies.

The DPPO objective for diffusion policy optimization is given by:
\begin{equation}
\resizebox{\linewidth}{!}{%
$\displaystyle
\mathcal{L}_{\text{DDPO}}(\theta) =
\mathbb{E}_{t} \left[
   \sum_{t=0}^{T}
   \min \big(
      w_t(\theta)\hat{A}_t,\,
      \text{clip}(w_t(\theta),\, 1-\epsilon,\, 1+\epsilon)\hat{A}_t
   \big)
\right]
$}
\end{equation}
\noindent 
where:
\begin{align}
w_t(\theta) &= 
\frac{p_\theta(\mathbf{x}_{t-1} \mid \mathbf{x}_t, \mathbf{c})}
     {p_{\theta_{\text{old}}}(\mathbf{x}_{t-1} \mid \mathbf{x}_t, \mathbf{c})}, \\
\hat{A}_t &= 
\nabla_\theta \log p_\theta(\mathbf{x}_{t-1} \mid \mathbf{x}_t, \mathbf{c}) \cdot r(\mathbf{x}_0, \mathbf{c}).
\end{align}

\noindent
Here, $w_t(\theta)$ is the importance weight that measures the likelihood ratio between the current and previous policies at denoising step $t$. The term $\hat{A}_t$ represents the advantage estimate, which measures how much better a particular denoising step is compared to the expected performance, and uses the final-step reward $r(\mathbf{x}_0, \mathbf{c})$ as a proxy for trajectory quality. The reward signal is sparse: only the final denoising step ($t=0$) receives the actual reward signal, which is then propagated backward through the entire denoising trajectory to assign credit. Over multiple iterations, this leads to a shift in the generative distribution toward motion outputs that better align with desired semantics, physical plausibility, or other downstream objectives.

Each training iteration follows a two-phase structure illustrated in Figure~\ref{fig:method}: \emph{Sample Collection} and \emph{Policy Update}. In the \emph{Sample Collection} phase, we construct a replay buffer by sampling prompts from the dataset and generating corresponding samples with the current DM. For each sample, we store the full diffusion trajectory, including all intermediate denoising steps, the sampled states $\mathbf{x}_{t-1}$, the likelihoods $p_{\theta_{\text{old}}}(\mathbf{x}_{t-1} \mid \mathbf{x}_t, \mathbf{c})$ and reward $r(\mathbf{x}_0, \mathbf{c})$.

In the \emph{Policy Update} phase, we train the DM using trajectories drawn from the replay buffer. We recompute the likelihoods of $x_{t-1}$ under the current model to perform importance sampling and update the parameters with the DDPO loss. This training is repeated for several epochs to fully exploit the collected data, after which the process restarts with a new \emph{Sample Collection} phase.

\input{Tables/cross}
\subsection{Reward Model}
\label{sec:Reward}
A key element of our approach is the reward model, responsible for accurately assessing how well a generated motion sequence matches a given textual prompt. Previous studies have investigated various models aimed at evaluating the quality of human motion generation~\cite{temos}, its consistency with human perception~\cite{reinforce_1}, and its alignment with language descriptions~\cite{Guo_2022_CVPR, tmr}.

Inspired by the success of post-training alignment techniques in other modalities using CLIP scores~\cite{clip}, which leverage cosine similarity between image and text embeddings in a shared semantic space, we adopt a similar strategy for the motion domain. This approach is particularly effective because cosine similarity in well-trained embedding spaces captures semantic alignment between modalities, allowing us to measure text-motion compatibility without requiring paired ground-truth data. 
Specifically, we employ a pretrained \textit{Text-Motion Retrieval} (TMR)~\cite{tmr} model as our reward function. 
The TMR model scores the compatibility between the generated motion $\mathbf{x}_0$ and the conditioning text $\mathbf{c}$, yielding a reward:
\begin{equation}
r(\mathbf{x}_0, \mathbf{c}) = \text{sim}(\phi_{\text{text}}(\mathbf{c}), \phi_{\text{motion}}(\mathbf{x}_0)),
\end{equation}

\noindent
where $\phi_{\text{text}}$ and $\phi_{\text{motion}}$ are text and motion encoders, and $\text{sim}(\cdot, \cdot)$ denotes cosine similarity. This score is computed between the prompt in input to the DM and the generated animation. This design enables reward computation without paired ground-truth data, making our method suitable for zero-shot generalization to new motion categories and styles.

\subsection{Efficient Learning}
\label{sec:EfficientLearning}
Reinforcement learning in DMs faces significant challenges due to sparse rewards, high-dimensional parameter spaces, and high computational demands. To address these issues, we incorporate two key strategies that improve both stability and efficiency: parameter-efficient fine-tuning via \textit{Low-Rank Adaptation} (LoRA)~\cite{hu2021lora}, and accelerated sampling with \textit{DPM-Solver++}~\cite{lu2025dpm}.

We stabilize RL fine-tuning using \textit{Low-Rank Adaptation} (LoRA)~\cite{hu2021lora}, a parameter-efficient fine-tuning technique that introduces low-rank trainable adapters into attention and MLP layers. We freeze the pretrained diffusion backbone and optimize only the LoRA layers. This reduces the number of trainable parameters and helps prevent overfitting, which is especially important when rewards are sparse or noisy.

To make this approach scalable, we replace the standard denoising process with \textit{DPM-Solver++}~\cite{lu2025dpm}, a high-order ODE-based sampler that significantly accelerates inference. While traditional diffusion sampling may require hundreds of steps, the DPM-Solver++ enables high-fidelity generation using as few as 10 steps. This dramatically reduces both memory and compute costs per training iteration, allowing for faster RL fine-tuning without sacrificing output quality.

\input{Figures/ominiclorati}
\section{Experiments}
%\subsection{Experimental Setup}
We evaluate our RL fine-tuning strategy on two diffusion-based motion generation models: StableMoFusion~\cite{huang2024stablemofusion} and MDM-SMPL~\cite{tevet2022humanmotiondiffusionmodel, petrovich24stmc}. StableMoFusion operates in a latent space using a latent diffusion framework and generates Guo-style motion features (\emph{guofeats})~\cite{Guo_2022_CVPR}, which represent human motion as a sequence of joint positions, velocities, and root trajectory information encoded in a compact feature space. In contrast, MDM-SMPL \cite{petrovich24stmc} is a model that directly generates motion in the SMPL format~\cite{loper2015smpl}, producing mesh parameters that define body shape and pose. These two models provide diversity in both representation and architecture, allowing us to validate the generalizability of our approach across different motion generation paradigms.

Our experiments were conducted on the HumanML3D~\cite{Guo_2022_CVPR} and KIT Motion-Language (KITML)~\cite{plappert2016kit} datasets. To assess the zero-shot generalization capability of our method, we perform a \textit{cross-dataset} evaluation: each model is pretrained on one dataset and then fine-tuned using only the textual prompts from the training set of the other dataset, without access to any motion ground truths.

In addition to the cross-dataset setting, we design a \textit{\nomemanipulation}~experiment on HumanML3D. We train a model on the full dataset excluding all samples from a specific action category, and then fine-tune it using our method only on the prompts from that held-out category. We define two such splits: \emph{Object Manipulation} (3,194 text-motion pairs involving object interactions) and \emph{Posture and Balance} (4,384 pairs related to seated or static postures). During the fine-tuning phase, for each split we further divide the prompts into training and evaluation subsets using an 80–20 ratio. Additional information about the \textit{\nomemanipulation}~settings are available in the supplementary materials.

During generation, we apply classifier-free guidance~\cite{ho2022classifier} with a scale of 2.5 for StableMoFusion and 5 for MDM-SMPL. These scales were chosen based on the values from the respective papers. Each model is fine-tuned for 30{,}000 iterations. At each iteration, we train for 4 epochs using a replay buffer containing 256 generated motion sequences. These sequences are produced by sampling 64 distinct prompts and replicating each prompt 4 times to enhance signal diversity.

For efficient and stable updates, we use LoRA~\cite{hu2021lora} with a rank of 4, scaling factor $\alpha = 16$, and no dropout. For sampling, we employ \textit{DPM-Solver++}~\cite{lu2025dpm} to reduce the number of denoising steps from 1000 to 10 for StableMoFusion and from 100 to 10 for MDM-SMPL. We did not use the Footskate cleanup optimization used in StableMoFusion due to its high computational cost and because such post-processing techniques are orthogonal to our contribution and can be applied independently to further improve results if desired.                                                                 
The evaluation metrics are: the Frechet Inception Distance (FID), which measures the distance between feature distributions of generated and real motions~\cite{Guo_2022_CVPR}; the Diversity metric, which quantifies motion variability through feature variance; MultiModality (MModality), which assesses the diversity of motions generated from the same text description; R Precision, which evaluates the accuracy of text-to-motion matching using Top-1 (R@1), Top-2 (R@2), and Top-3 (R@3) retrieval accuracy~\cite{Guo_2022_CVPR}; and MultiModal Distance (MMDist), that represent the distance between the representations of the generated motion and the prompt.

\input{Tables/leave}
\subsection{Cross-Dataset Evaluation}
In Table~\ref{tab:cross_kit} we report results on cross-dataset experiments in which models are trained on HumanML3D and evaluated on the KIT Motion-Language test set (Human-to-Kit). Table~\ref{tab:cross_humanml} presents results from the opposite setting (Kit-to-Human). Results are consistent in both directions.

We observe that pretrained models suffer a strong performance drop in cross-dataset evaluation compared to their in-domain results, confirming the limited generalization ability of current approaches. Notably, the FID is particularly high in the cross-dataset setting, while retrieval scores remain relatively stable. This suggests that while models trained on a different dataset can still generate semantically relevant motions, these motions deviate significantly from the target distribution's stylistic and kinematic characteristics, leading to poor FID values. The effect is most pronounced in the Kit-to-Human setting, which is further hindered by the smaller size of the KIT-ML training set, containing about 3,900 sequences compared to HumanML3D's 14,600.

Our RL fine-tuning approach effectively addresses these limitations. For our models, we observe consistent improvements in both retrieval scores and FID when fine-tuning MDM-SMPL and StableMoFusion, achieving the best performance among all models. Specifically, FID improvements range from 15-30\% across settings, while retrieval scores improve by 2-5\%. In contrast, MultiModality slightly decreases, suggesting that our RL-based approach prioritizes semantic alignment with over generating a wide range of motion variations. The Diversity metric remains largely stable across both pretrained and fine-tuned models.

It is worth noting that FID is substantially lower for MDM-SMPL largely because the SMPL representation restricts the model’s expressivity. While this limits generation diversity, it also forces outputs to remain closer to the ground-truth distribution, as the parameterization inherently constrains the space of possible motions to anatomically plausible configurations.

Examples of generated motions are shown in Figure~\ref{fig:comparison}. These highlight how RL fine-tuning improves accuracy. For instance, in examples (b) and (c), the pretrained model confuses clockwise with counterclockwise movements, while the fine-tuned model demonstrates greater robustness. Similarly, in example (a) and (d), the fine-tuned model better follows the input description showing more expressive animations.

\input{Figures/userstudy}

\input{Tables/forgetting}
This stronger text adherence can be attributed to the difference in training objectives. Pretraining with an MSE loss struggles to capture subtle distinctions, such as left versus right or clockwise versus counterclockwise, because these prompts have very similar text embeddings in the CLIP text encoder space, which makes them difficult to distinguish during standard diffusion training. Our reward model (TMR), instead, is optimized with a contrastive loss designed to separate such closely related concepts in the embedding space, thus allowing the model to follow instructions more precisely.

\subsection{User Study}
To complement the quantitative evaluation, we conducted a user study in the Human-to-Kit setting using an A/B testing protocol. Thirty participants, including both motion analysis experts and general users, compared a total of 20 pairs of motions generated by our fine-tuned StableMoFusion and by the four pretrained baseline approaches. Each pair was evaluated along two dimensions: (i) overall realism; (ii) adherence to the textual prompt.

In Figure~\ref{fig:userstudy}, we show the results of the study. Our fine-tuned model outperforms the baselines in both text adherence and motion realism. While the advantage is more modest over StableMoFusion and MotionGPT, the preference for our method becomes more pronounced when compared to MoMask and MDM-SMPL.

\subsection{\nomemanipulation~Experiments}
In Table~\ref{tab:leaveoneout}, we report results from our \nomemanipulation~scenario. In this setting, the StableMoFusion model is first trained on a subset of HumanML3D with one motion class removed. We then fine-tune the model on the held-out class using our approach and evaluate on a test set from that class. This setup is designed to assess the effectiveness of our method in adapting models to previously unseen motion categories. For reference, we also include the performance of the original StableMoFusion model trained on the full dataset (`Full model'), evaluated on the same test sets.

In the \emph{Object Manipulation} experiment (Table~\ref{tab:leaveoneout}(a)), our approach improves both retrieval scores and FID, even surpassing the result of the model trained on the full dataset. As observed in other experiments, MultiModality is slightly reduced in favor of stronger semantic alignment.

In the \emph{Posture and Balance}  experiment (Table~\ref{tab:leaveoneout}(b)), the results follow the same pattern as the first experiment with an even higher improvement on the full dataset model performances, confirming the effectiveness of our method across different motion categories.

\subsection{Forgetting}
To assess the impact of our post-training approach on performance over the original data distribution, we evaluate the fine-tuned models on the test sets of their pretraining datasets. Table~\ref{tab:forgetting} reports the results for the Human-to-Kit cross-dataset experiment (additional results for other settings are included in the Supplementary Material). Remarkably, performance does not degrade after fine-tuning: both retrieval scores and FID even show slight improvements, suggesting the presence of positive backward transfer between datasets. This indicates that exposure to different motion styles and descriptions during RL fine-tuning actually enhances the model's understanding of motion-text relationships, leading to improved performance even on the original dataset. Consistent with previous experiments, we observe a slight reduction in the MultiModality metric, indicating a shift toward stronger semantic alignment at the expense of some variability in the generated motions.

\section{Conclusions}
We presented an RL-based post-training framework for adapting human motion diffusion models to new datasets and unseen motion categories without requiring additional motion capture data. Our method leverages a pretrained text-motion retrieval model as the sole reward signal, enabling ground truth-free fine-tuning that is both data-efficient and privacy-preserving.

Through extensive experiments, we demonstrated that our approach consistently improves retrieval scores and FID in cross-dataset and \nomemanipulation~settings, closing the gap with fully trained models. Importantly, we observed that fine-tuned models maintain performance on the original data distribution, with slight improvements, indicating positive backward transfer. The main trade-off is a modest reduction in MultiModality, as the model prioritizes semantic alignment with textual prompts over the diversity of motions generated from the same description.

Overall, our findings highlight the potential of RL as a practical tool for post-training alignment of motion DMs. By eliminating the need for costly motion capture data and full retraining, our framework offers a scalable path toward more adaptable and deployable human motion generation systems. 

{
    \small
    \bibliographystyle{ieeenat_fullname}
    \bibliography{main}
}

\end{document}

%% file: Figures/teaser.tex
\begin{figure*}[htbp]
\centering
\includegraphics[width=\textwidth]{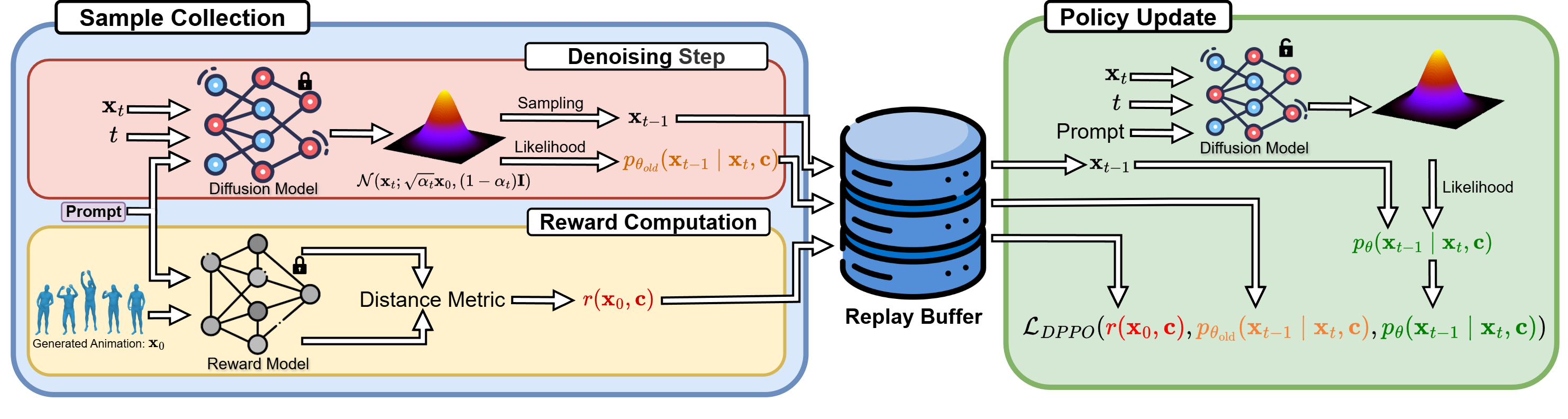} 
\caption{Overview of our fine-tuning procedure. \textbf{Left: Sample Collection.} Diffusion trajectories are generated from Gaussian noise conditioned on prompts sampled from the dataset. At each denoising step, the model outputs a normal distribution from which $\mathbf{x}_{t-1}$ is sampled; the sample and its likelihood $p_{\theta}(\mathbf{x}_{t-1} \mid \mathbf{x}_t, \mathbf{c})$, along with the timestep, input, and prompt, are stored in the replay buffer. After denoising, the final animation is evaluated by the reward model, which embeds both the prompt and the animation into a joint space and assigns a reward based on their embedding distance. \textbf{Right: Policy Update.} Trajectories are sampled from the replay buffer, likelihoods are recomputed with the current DM, and the model is updated using the DDPO loss.}
\label{fig:method}
\end{figure*}

%% file: Tables/cross.tex
\begin{table*}[ht]
\caption{Cross-Dataset Results. The base model is pretrained on HumanML3D and evaluated on KIT-ML in (a), while in (b), the model is pretrained on KIT-ML and evaluated on HumanML3D. We compare zero-shot approaches and post-training with our method, which fine-tunes the model without relying on ground-truth annotations.}
\centering

% ---------- Subtable 1 ----------
\begin{subtable}{\textwidth}
\centering
\caption{Train on HumanML3D, test on KIT-ML}
\begin{tabular}{lcccccccc}
\toprule
\textbf{Method} & \textbf{R@1 ↑} & \textbf{R@2 ↑} & \textbf{R@3 ↑} & \textbf{FID ↓} & \textbf{MMDist ↓} & \textbf{Diversity →} & \textbf{MModality ↑} \\
\midrule
Ground Truth & 0.401 & 0.601 & 0.730 & 0 & 2.636 & 9.103 & -- \\
\midrule
MoMask & 0.385 & 0.574 & 0.688 & 1.622 & 2.994 & \textbf{9.058} & 1.198 \\
MotionGPT & 0.368 & 0.552 & 0.651 & 2.740 & 3.721 & 8.845 & 2.342 \\
StableMoFusion & 0.362 & 0.553 & 0.664 & 1.860 & 3.104 & 8.603 & \textbf{2.497} \\
MDM-SMPL & 0.257 & 0.412 & 0.530 & 0.920 & 3.146 & 9.308 & 1.024 \\
\midrule
\rowcolor{tabhighlight}StableMoFusion (ours) & \textbf{0.413} & \textbf{0.618} & \textbf{0.732} & 1.291 & \textbf{2.830} & 8.730 & 1.812 \\
\rowcolor{tabhighlight}MDM-SMPL (ours) & 0.261 & 0.398 & 0.522 & \textbf{0.614} & 3.119 & 9.267 & 0.926 \\
\bottomrule
\end{tabular}
\label{tab:cross_kit}
\end{subtable}
\\[1em]
% ---------- Subtable 12----------
\begin{subtable}{\textwidth}
\centering
\caption{Train on KIT-ML, test on HumanML3D}
%\begin{table*}[ht]
%\caption{Cross-dataset results on HumanML3D. The base model is pretrained on KIT-ML and evaluated on HumanML3D in a zero-shot setting or after post-training with our approach, which finetunes the model without using ground-truth annotations.}
%\centering
%\small
\begin{tabular}{lcccccccc}
\toprule
\textbf{Method} & \textbf{R@1 ↑} & \textbf{R@2 ↑} & \textbf{R@3 ↑} & \textbf{FID ↓} & \textbf{MMDist ↓} & \textbf{Diversity →} & \textbf{MModality ↑} \\
\midrule
Ground Truth & 0.518 & 0.709 & 0.807 & 0 & 2.956 & 9.649 & -- \\
\midrule
MoMask & 0.337 & 0.513 & 0.645 & 1.923 & 3.410 & \textbf{9.536} & 1.142 \\
MotionGPT & 0.231 & 0.347 & 0.437 & 5.018 & 5.954 & 9.805 & \textbf{2.129} \\
StableMoFusion & 0.327 & 0.488 & 0.589 & 2.465 & 4.355 & 8.424 & 1.210 \\
MDM-SMPL & 0.276 & 0.428 & 0.531 & 1.368 & 4.705 & 9.297 & 1.920 \\
\midrule
\rowcolor{tabhighlight}StableMoFusion (ours) & \textbf{0.391} & \textbf{0.594} & \textbf{0.711} & 1.799 & \textbf{3.263} & 8.833 & 1.194 \\
\rowcolor{tabhighlight}MDM-SMPL (ours) & 0.283 & 0.431 & 0.542 & \textbf{0.975} &  4.561 & 9.112 & 1.824 \\
\bottomrule
\end{tabular}
\label{tab:cross_humanml}
\end{subtable}
\end{table*}

%% file: Figures/ominiclorati.tex
\begin{figure*}[ht]
\centering
\renewcommand{\arraystretch}{1.3} % increase row height
\setlength{\tabcolsep}{4pt} % column spacing
\resizebox{\textwidth}{!}{%
\begin{tabular}{>{\centering\arraybackslash}m{0.2cm}  
                >{\centering\arraybackslash}m{3.5cm} 
                >{\centering\arraybackslash}m{3.5cm} 
                >{\centering\arraybackslash}m{3.5cm} 
                >{\centering\arraybackslash}m{3.5cm}} 

    & \textbf{(a)} A person raises both arms like jumping jack. 
    & \textbf{(b)} A person walks in a circle clockwise. 
    & \textbf{(c)} A person walks counter-clockwise in a circle. 
    & \textbf{(d)} A person waves his right hand. \\

    \rotatebox{90}{\textbf{Pretrained}} &
    \includegraphics[width=0.9\linewidth]{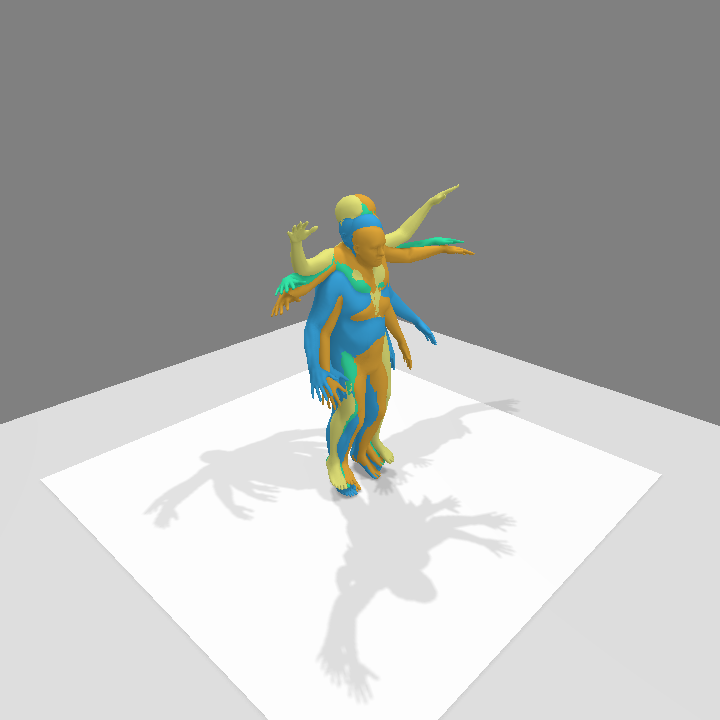} &
    \includegraphics[width=0.9\linewidth]{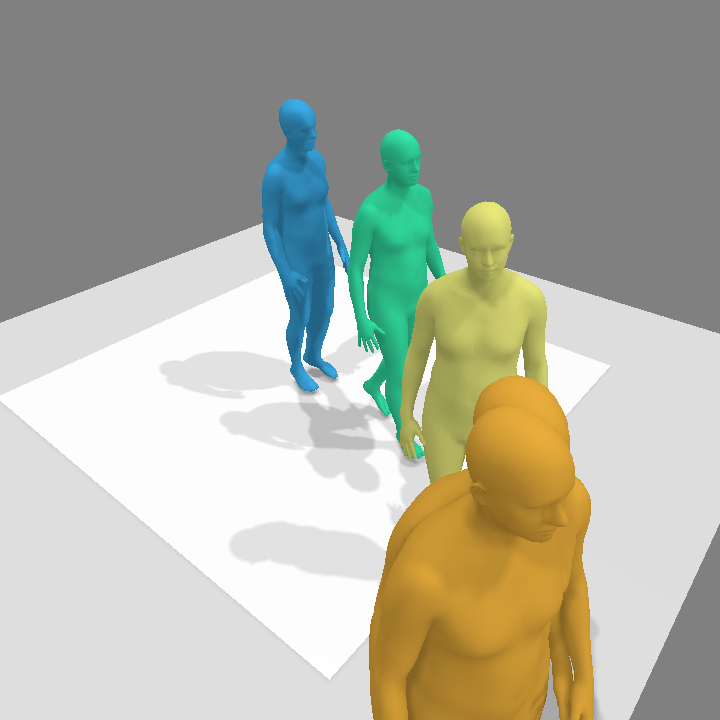} &
    \includegraphics[width=0.9\linewidth]{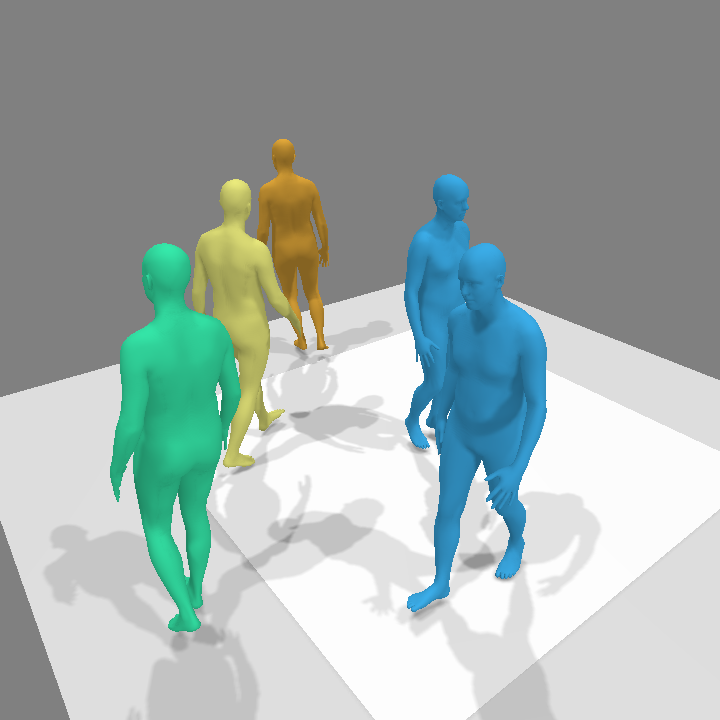} &
    \includegraphics[width=0.9\linewidth]{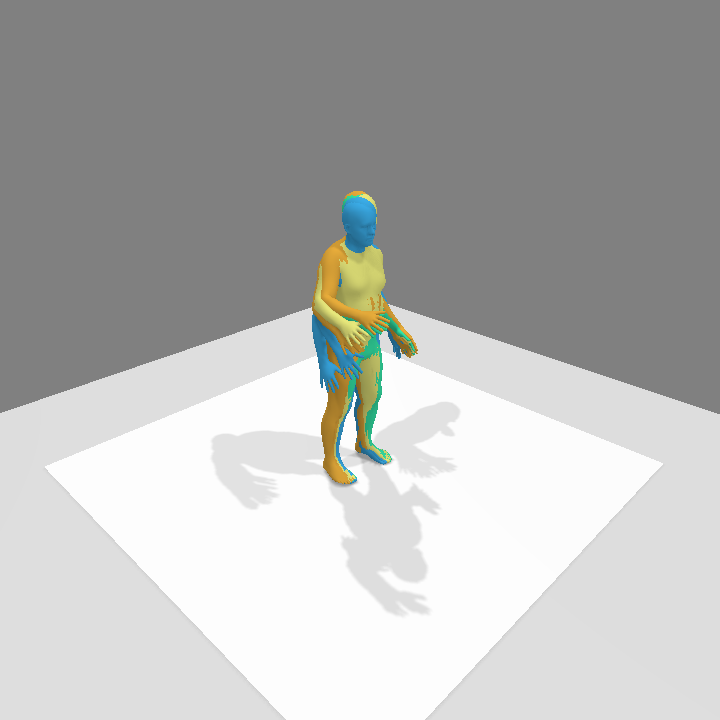} \\

    \rotatebox{90}{\textbf{Ours}} &
    \includegraphics[width=0.9\linewidth]{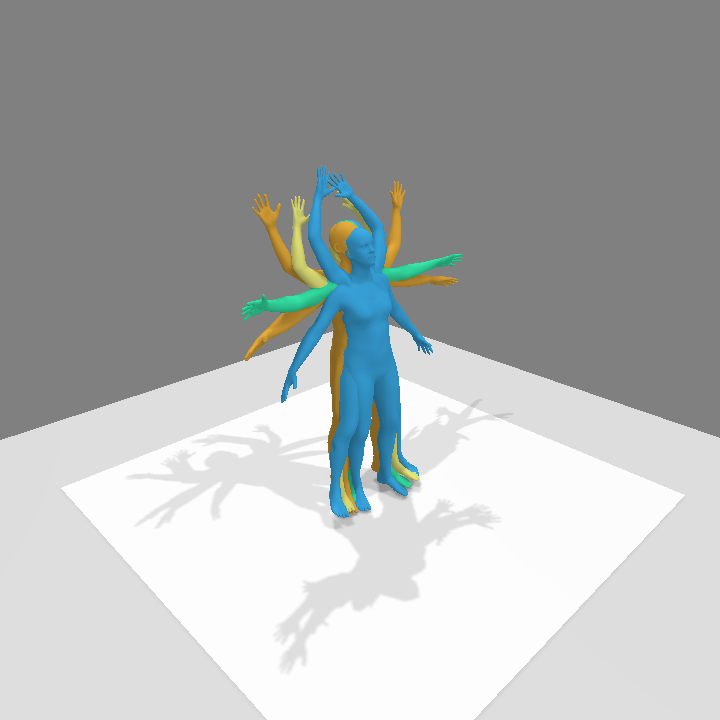} &
    \includegraphics[width=0.9\linewidth]{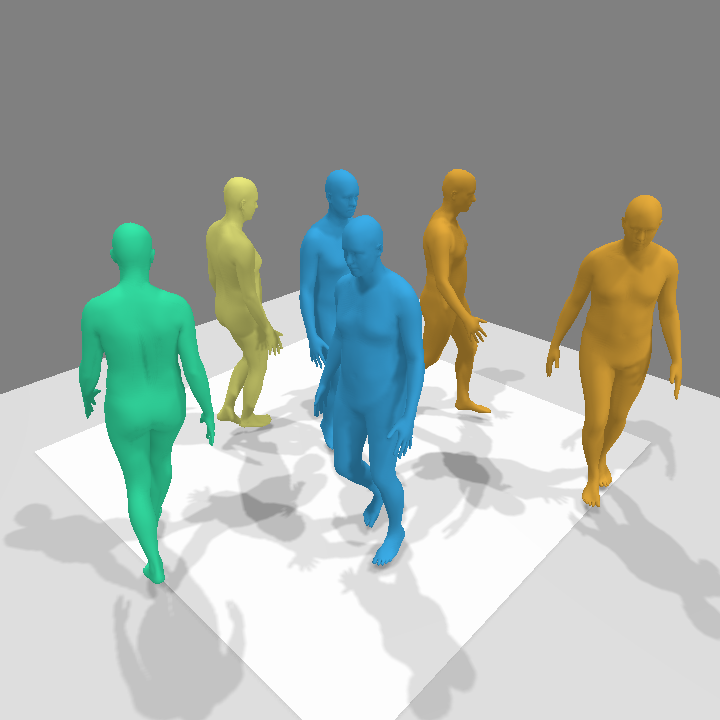} &
    \includegraphics[width=0.9\linewidth]{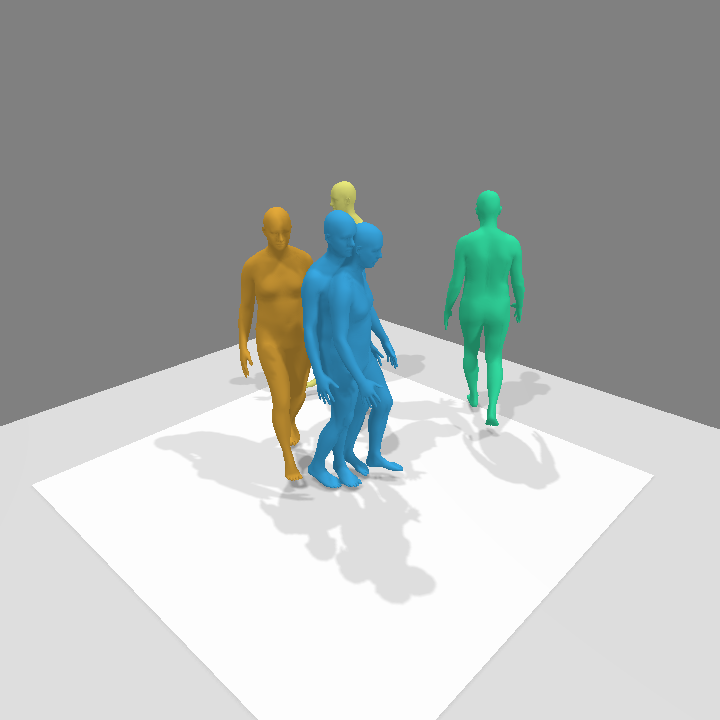} &
    \includegraphics[width=0.9\linewidth]{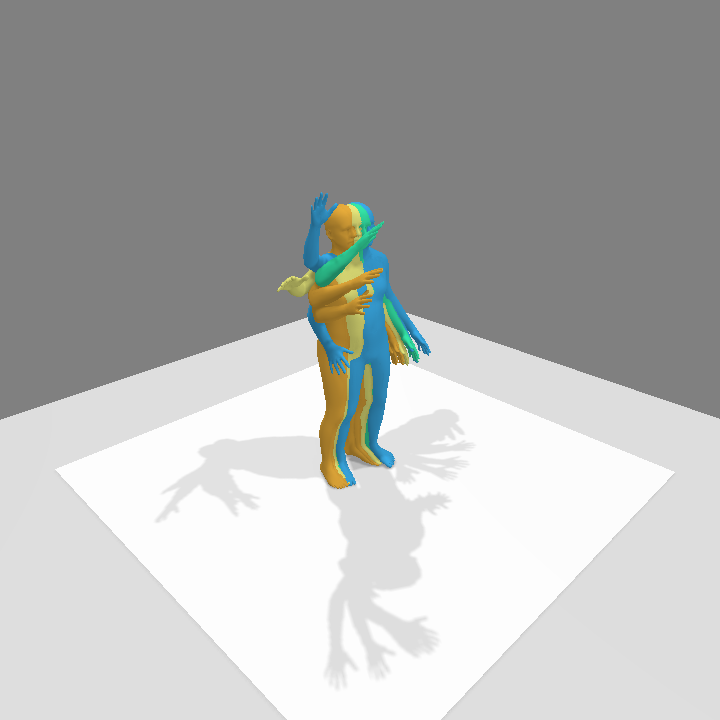} \\

\end{tabular}
}
\caption{Example of improved text adherence after our fine-tuning of the StableMoFusion model. The figure shows the full animation, with color indicating time from blue to orange. The first row depicts the model before fine-tuning, while the second row shows the model after fine-tuning. After fine-tuning, the generated motions better follow the textual prompts. In particular, in panels (b) and (c), the model fully completes the circular motion, and in panels (a) and (b), the hand movements are more expressive.}
\label{fig:comparison}
\end{figure*}

%% file: Tables/leave.tex
\begin{table*}[ht]
\centering

\caption{\nomemanipulation { }results on HumanML3D. A model is trained from scratch with one motion class removed from the dataset, fine-tuned using our approach, and then evaluated on a test set containing only the held-out class.}
\label{tab:leaveoneout}
% ---------- Subtable 1 ----------
\begin{subtable}{\textwidth}
\centering
\caption{Object Manipulation}
\begin{tabular}{lcccccccc}
\toprule
\textbf{Method} & \textbf{R@1 ↑} & \textbf{R@2 ↑} & \textbf{R@3 ↑} & \textbf{FID ↓} & \textbf{MMDist ↓} & \textbf{Diversity →} & \textbf{MModality ↑} \\
\midrule
Ground Truth & 0.403 & 0.585 & 0.700 & 0     &  2.617 & 8.077 & -- \\
Full model   & 0.344 & 0.529 & 0.641 & 0.584 & 2.973  & 7.375 & 1.896 \\
\midrule
StableMoFusion & 0.331 & 0.509 & 0.617 & 0.714 &  3.121 & 7.442 & \textbf{1.832} \\
\rowcolor{tabhighlight}StableMoFusion (ours) & \textbf{0.351} & \textbf{0.528} & \textbf{0.639} & \textbf{0.615} &  \textbf{2.939} & \textbf{7.626} & 1.804 \\
\bottomrule
\end{tabular}
\end{subtable}
\\[1em]
% ---------- Subtable 2 ----------
\begin{subtable}{\textwidth}
\centering
\caption{Posture and Balance}
\begin{tabular}{lcccccccc}
\toprule
\textbf{Method} & \textbf{R@1 ↑} & \textbf{R@2 ↑} & \textbf{R@3 ↑} & \textbf{FID ↓} & \textbf{MMDist ↓} & \textbf{Diversity →} & \textbf{MModality ↑} \\
\midrule
Ground Truth & 0.475 & 0.652 & 0.761 & 0     &  2.530 & 8.217 & -- \\
Full model   & 0.383 & 0.577 & 0.691 & 0.400 & 2.893  & 7.099 & 1.719 \\
\midrule
StableMoFusion & 0.378 & 0.563 & 0.668 & 0.432 &  2.930 & 7.082 & \textbf{1.554} \\
\rowcolor{tabhighlight}StableMoFusion (ours) & \textbf{0.424} & \textbf{0.608} & \textbf{0.720} & \textbf{0.335} &  \textbf{2.734} & \textbf{7.553} & 1.468 \\
\bottomrule
\end{tabular}
\end{subtable}
\end{table*}

%% file: Figures/userstudy.tex
\begin{figure}[!ht]
    \centering
    \includegraphics[width=0.45\textwidth]{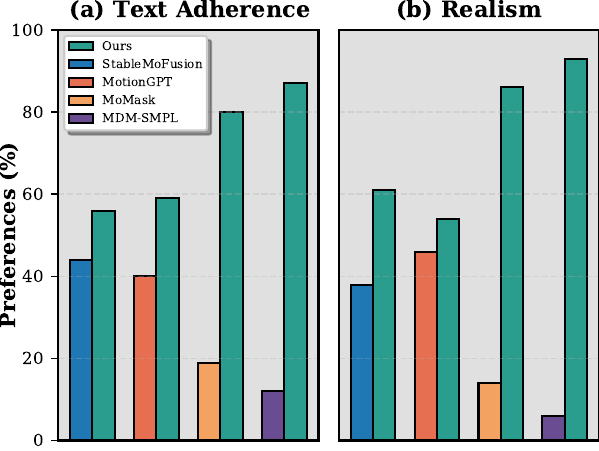} 
    \caption{\textbf{Perception study results}: Human raters evaluated our method against pretrained baseline models in the Human-to-Kit scenario, assessing both motion realism and text adherence in an A/B scenario.}
    \label{fig:userstudy}
\end{figure}

%% file: Tables/forgetting.tex
\begin{table*}[ht]
\caption{Ablation study on forgetting after fine-tuning. We evaluate models pretrained on HumanML3D and fine-tuned on KIT-ML, reporting results on the HumanML3D test set to assess the impact of fine-tuning on the original distribution. The results show no performance degradation and, even improvements, indicating backward transfer.}
\centering

\begin{tabular}{lcccccccc}
\toprule
\textbf{Method} & \textbf{R@1 ↑} & \textbf{R@2 ↑} & \textbf{R@3 ↑} & \textbf{FID ↓} & \textbf{MMDist ↓} & \textbf{Diversity →} & \textbf{MModality ↑} \\
\midrule
Ground Truth & 0.518 & 0.709 & 0.807 & 0    & 2.956  & 9.320 & -- \\
\hline
MoMask       & 0.521 & 0.713 & 0.807 & 0.045 &  2.958 & --    & 1.241 \\
MotionGPT    & 0.492 &   0.681  &  0.778  & 0.232 & 3.096 & 9.528 & -- \\
StableMoFusion & 0.492 & 0.686 & 0.787 & 0.500 & 3.104 & 8.876 & 1.955 \\
MDM-SMPL     & 0.395 & 0.574 & 0.678 & 0.380 & 3.866 & 9.255 & 1.313 \\
\midrule
\rowcolor{tabhighlight}StableMoFusion (Ours) & 0.502 & 0.696 & 0.796 & 0.400 & 3.053  & 8.984 & 1.852 \\
\rowcolor{tabhighlight}MDM-SMPL (Ours) & 0.400 & 0.577 & 0.682 & 0.373 & 3.817 & 9.207 & 1.312 \\
\bottomrule
\end{tabular}
\label{tab:forgetting}
\end{table*}